\documentclass[10pt,twocolumn,letterpaper]{article}
\usepackage[pagenumbers]{cvpr} 

\usepackage{graphicx}
\usepackage{amsmath}
\usepackage{amssymb}
\usepackage{booktabs}
\usepackage{multirow}
\usepackage{algorithm}  
\usepackage{algorithmicx}
\usepackage{algpseudocode}
\usepackage{float}
\usepackage[symbol]{footmisc}

\usepackage[pagebackref,breaklinks,colorlinks]{hyperref}
\newlength\savewidth\newcommand\shline{\noalign{\global\savewidth\arrayrulewidth
  \global\arrayrulewidth 1pt}\hline\noalign{\global\arrayrulewidth\savewidth}}

\usepackage[capitalize]{cleveref}
\crefname{section}{Sec.}{Secs.}
\Crefname{section}{Section}{Sections}
\Crefname{table}{Table}{Tables}
\crefname{table}{Tab.}{Tabs.}

\def\0{{\bf 0}}
\def\1{{\bf 1}}

\usepackage{xcolor}
\definecolor{mygreen}{RGB}{10,186,181}

\definecolor{citecolor}{HTML}{0071bc}
\definecolor{paleplum}{rgb}{0.8, 0.6, 0.8}
\hypersetup{
  colorlinks,
  citecolor=violet,
  linkcolor=red
}

 \makeatletter
\def\@fnsymbol#1{\ensuremath{\ifcase#1\or \dagger\or \ddagger\or
   \mathsection\or \mathparagraph\or \|\or **\or \dagger\dagger
   \or \ddagger\ddagger \else\@ctrerr\fi}}
\makeatother

\begin{document}

\title{Stitchable Neural Networks}

\author{Zizheng Pan  \quad Jianfei Cai \quad Bohan Zhuang\footnotemark  \\ [0.2cm]
ZIP Lab, Monash University \\ [0.2cm]
\url{https://snnet.github.io}
}

% \maketitle
\twocolumn[{
\maketitle
\vspace{-10mm}
\begin{figure}[H]
\hsize=\textwidth
\centering
    \includegraphics[width=0.95\textwidth]{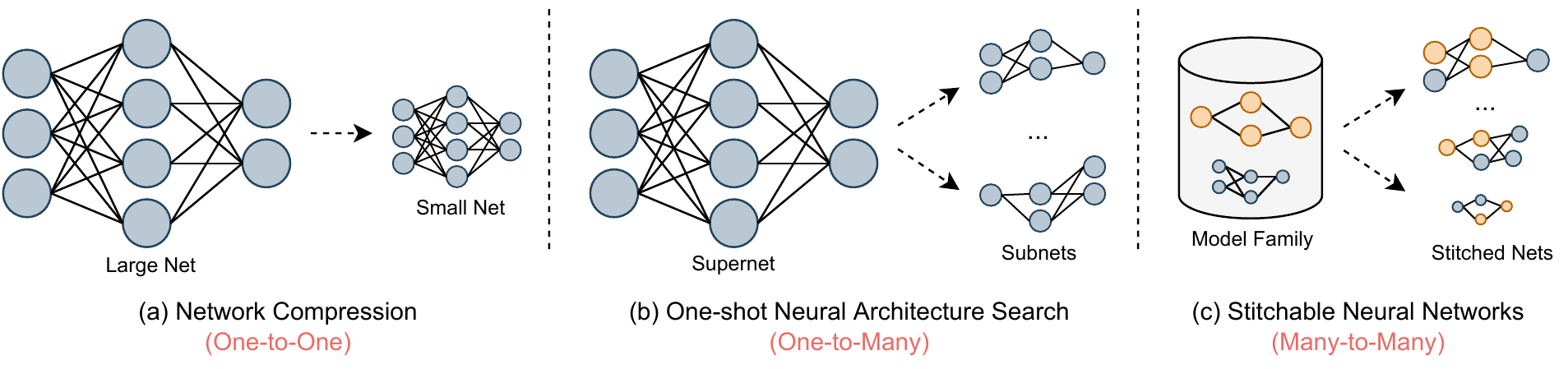}
    \caption{Compared with previous scalable deep learning frameworks. (a) Network compression shrinks a large network into a small one by techniques such as pruning, quantization and knowledge distillation, \etc, which is a one-to-one mapping.
    (b) One-shot neural architecture search first trains a supernet that supports diverse architectural settings and then specializes a subnet given the target resource constraint during deployment, which is a case of one-to-many. (c) Our proposed Stitchable Neural Network directly stitches the off-the-rack family of pretrained models and quickly obtains new networks for efficient model design and deployment in a novel many-to-many paradigm.}
    \label{fig:banner}	  
\end{figure}
}]

\footnotetext[1]{Corresponding author. E-mail: $\tt  bohan.zhuang@gmail.com$}

%%%%%%%%% ABSTRACT
\begin{abstract}
The public model zoo containing enormous powerful pretrained model families (\eg, ResNet/DeiT) has reached an unprecedented scope than ever, which significantly contributes to the success of deep learning. As each model family consists of pretrained models with diverse scales (\eg, DeiT-Ti/S/B), it naturally arises a fundamental question of how to efficiently assemble these readily available models in a family for dynamic accuracy-efficiency trade-offs at runtime. To this end, we present Stitchable Neural Networks (SN-Net), a novel scalable and efficient framework for model deployment. It cheaply produces numerous networks with different complexity and performance trade-offs given a family of pretrained neural networks, which we call anchors. Specifically, SN-Net splits the anchors across the blocks/layers and then stitches them together with simple stitching layers to map the activations from one anchor to another. With only a few epochs of training, SN-Net effectively interpolates between the performance of anchors with varying scales. At runtime, SN-Net can instantly adapt to dynamic resource constraints by switching the stitching positions. 
Extensive experiments on ImageNet classification demonstrate that SN-Net can obtain on-par or even better performance than many individually trained networks while supporting diverse deployment scenarios. 
For example, by stitching Swin Transformers, we challenge hundreds of models in Timm model zoo with a single network. We believe this new elastic model framework can serve as a strong baseline for further research in wider communities.
\end{abstract}

%%%%%%%%% BODY TEXT
\section{Introduction}
\label{sec:intro}
The vast computational resources available and large amount of data have driven researchers to build tens of thousands of powerful deep neural networks with strong performance, which have largely underpinned the most recent breakthroughs in machine learning and much broader artificial intelligence.
Up to now, there are $\sim$81k models on HuggingFace~\cite{wolf-etal-2020-transformers} and $\sim$800 models on Timm~\cite{rw2019timm} that are ready to be downloaded and executed without the overhead of reproducing. Despite the large model zoo, a model family (\eg, DeiT-Ti/S/B~\cite{deit}) that contains pretrained models with functionally similar architectures but different scales only covers a coarse-grained level of model complexity/performance, where each model only targets a specific resource budget (\eg, FLOPs). Moreover, the model family is not flexible to adapt to dynamic resource constraints since each individual model is not re-configurable due to the fixed computational graph. In reality, we usually need to deploy models to diverse platforms with different resource constraints (\eg, energy, latency, on-chip memory). For instance, a mobile app in Google Play has to support tens of thousands of unique Android devices, from a high-end Samsung Galaxy S22 to a low-end Nokia X5. Therefore, given a family of pretrained models in the model zoo, a fundamental research question naturally arises: \emph{how to effectively utilise these off-the-shelf pretrained models to handle diverse deployment scenarios for Green AI~\cite{greenai}?}

\begin{figure}[]
	\centering
	\includegraphics[width=1.0\linewidth]{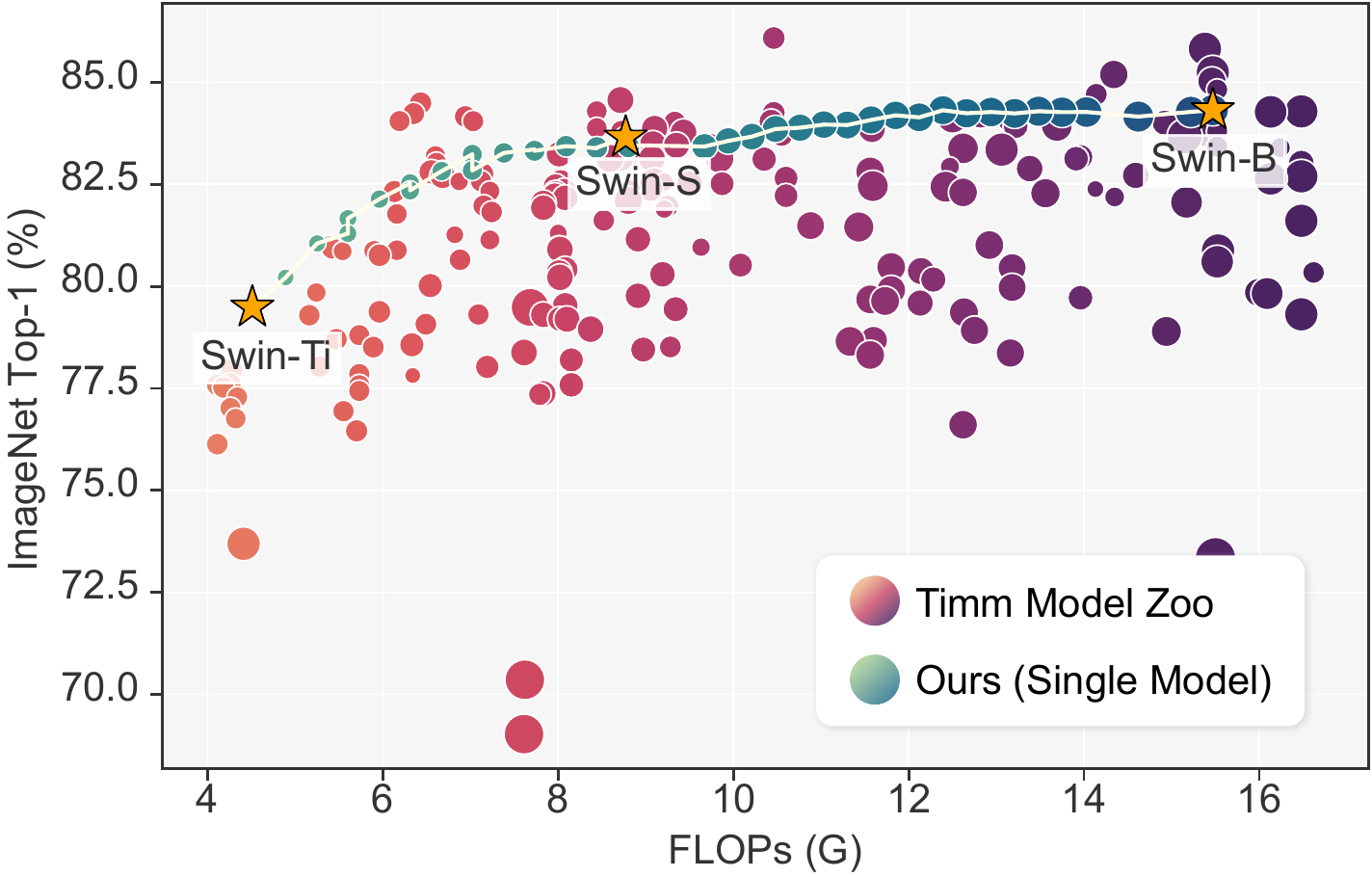}
	\caption{\textbf{One} Stitchable Neural Network \vs \textbf{200} models in Timm model zoo~\cite{rw2019timm}. It shows an example of SN-Net by stitching ImageNet-22K pretrained Swin-Ti/S/B. Compared to each individual network, SN-Net is able to instantly switch network topology at runtime and covers a wide range of computing resource budgets. Larger and darker dots indicate a larger model with more parameters and higher complexity.}
	\vspace{-20pt}
	\label{fig:compare_model_zoo}
\end{figure}

To answer this question, a naive solution is to train individual models with different accuracy-efficiency trade-offs from scratch. However, such method has a linearly increased training and time cost with the number of possible cases. Therefore, one may consider the existing scalable deep learning frameworks, such as model compression and neural architecture search (NAS), to obtain models at different scales for diverse deployment requirements. Specifically, network compression approaches such as pruning~\cite{0022KDSG17,snip,LinJWZZ0020}, quantization~\cite{xnor_net,lq_nets,data_free_quant} and knowledge distillation~\cite{fitnets,SunCGL19,Chen0ZJ21} aim to obtain a small model from a large and well-trained network, which however only target one specific resource budget (see Figure \ref{fig:banner} (a)), thus not flexible to meet the requirements of real-world deployment scenarios. On the other hand, one-shot NAS~\cite{PhamGZLD18,darts}, a typical NAS framework that decouples training and specialization stages, seeks to train an over-parameterized supernet that supports many sub-networks for run-time dynamics (see Figure \ref{fig:banner} (b)), but training the supernet is extremely time-consuming and computationally expensive (\eg, 1,200 GPU hours on 32 V100 GPUs in OFA~\cite{ofa}). To summarize, the existing scalable deep learning frameworks are still limited within a single model design space, which cannot inherit the rich knowledge from pretrained model families in a model zoo for better flexibility and accuracy. Besides, they also require complicated training strategies to guarantee a good model performance.

In this work, we present Stitchable Neural Network (SN-Net), a novel scalable deep learning framework
for efficient model design and deployment which quickly stitches an off-the-shelf pretrained model family with much less training effort to cover a fine-grained level of model complexity/performance for a wide range of deployment scenarios (see Figure~\ref{fig:banner} (c)). Specifically, SN-Net is motivated by the previous observations~\cite{stitch_0,stitch_1,stitch_2} that the typical minima reached by SGD can be stitched to each other with low loss penalty, which implies architectures of the same model family pretrained on the same task can be stitched. Based on this insight, SN-Net directly selects the well-performed pretrained models in a model family as ``anchors'', and then inserts a few simple stitching layers at different 
positions to transform the activations from one anchor to its nearest anchor in terms of complexity. In this way, SN-Net naturally interpolates a path between neighbouring anchors of different accuracy-efficiency trade-offs, and thus can handle dynamic resource constraints \textit{with a single neural network at runtime}. An example is shown in Figure~\ref{fig:compare_model_zoo}, where a single Swin-based SN-Net is able to do what hundreds of models can do with only 50 epochs training on ImageNet-1K.

We systematically study the design principles for SN-Net, including the choice of anchors, the design of stitching layers, the stitching direction and strategy, along with a sufficiently simple but effective training strategy. With comprehensive experiments, we show that SN-Net demonstrates promising advantages: 1) Compared to the existing prevalent scalable deep learning frameworks (Figure~\ref{fig:banner}), SN-Net is a new universal paradigm which breaks the limit of a single pretrained model or supernet design by extending the design space into a large number of model families in the model zoo, forming a ``many-to-many'' pipeline. 2) Different from NAS training that requires complex optimization techniques~\cite{ofa,slimmable_nn_v1}, training SN-Net is as easy as training individual models while getting rid of the huge computational cost of training from scratch. 3) The final performance of stitches is almost predictable due to the interpolation-like performance curve between anchors, which implies that we can selectively train a number of stitches prior to training based on different deployment scenarios.

In a nutshell, we summarize our contributions as follows:
\begin{itemize}
\itemsep -0.05cm
  \item We introduce Stitchable Neural Networks, a new universal framework for elastic deep learning
  by directly utilising the pretrained model families in  model zoo via model stitching.
  \item We provide practical principles to design and train SN-Net, laying down the foundations for future research.
  \item Extensive experiments demonstrate that compared to training individual networks from scratch, \eg, a single DeiT-based~\cite{deit} SN-Net can achieve flexible accuracy-efficiency trade-offs at runtime while reducing $22\times$ training cost and local disk storage.
\end{itemize}
\section{Related Work} \label{sec:related_work}

\paragraph{Model stitching.}
Model stitching was initially proposed by Lenc~\etal~\cite{stitch_0} to study the equivalence of representations. Specifically, they showed that the early portion of a trained network can be connected with the last portion of another trained network by a $1\times1$ convolution stitching layer without significant performance drop. Most recently, Yamini~\etal~\cite{stitch_1} revealed that neural networks, even with different architectures or trained with different strategies, can also be stitched together with small effect on performance. As a concurrent work to \cite{stitch_1}, Adrián~\etal~\cite{stitch_2} studied using model stitching as an experimental tool to match neural network representations. They demonstrated that common similarity indices (\eg, CKA~\cite{cka}, CCA~\cite{cca}, SVCCA~\cite{svcca}) are not correlated to the performance of the stitched model. Unlike these previous works which view model stitching as a tool to measure neural network representations, this paper unleashes the power of model stitching as a general approach for utilising the pretrained model families in the large-scale model zoo to obtain a single scalable neural network at a low cost that can instantly adapt to diverse deployment scenarios. More recently, Yang~\etal proposed DeRy~\cite{dmr} to dissect and reassemble arbitrary pretrained models into a new network for a certain resource constraint (\eg, FLOPs) one at a time. Unlike DeRy, the proposed SN-Net supports numerous sub-networks by stitching the off-the-shelf model families, being capable of handling diverse resource budgets at deployment time.

\paragraph{Neural architecture search.}
Neural architecture search (NAS)~\cite{ZophL17} aims to automatically search for the well-performed network architecture in a pre-defined search space under different resource constraints. In the early attempts~\cite{ZophL17,ZophVSL18}, NAS consumes prohibitive computational cost (\eg, 500 GPUs across 4 days in~\cite{ZophVSL18}) due the requirement of training individual sub-networks until convergence for accurate performance estimation. To address this problem, one-shot NAS~\cite{PhamGZLD18,darts,LuoTQCL18,CaiZH19,XieZLL19}  has been proposed to improve NAS efficiency by weight sharing, where multiple subnets share the same weights with the supernet. However, training a supernet still requires intensive computing resources. Most recently, zero-shot NAS~\cite{IstrateSMNBM19,AbdelfattahMDL21,MellorTSC21,ChenGW21} has been proposed to identify good architectures prior to training. However, obtaining the final model still requires training from scratch. Compared to NAS, our method builds upon the off-the-shelf  family of pretrained models in model zoo, which exploits the large model design space and is able to assemble the existing rich knowledge from heterogeneous models for flexible and diverse model deployments.

\paragraph{Vision Transformers.}
Vision Transformers~\cite{vit} are emerging deep neural networks which have challenged the de-facto standard of convolutional neural networks on vision tasks. The majority of the existing efforts focus on improving the performance of ViT as a single general vision backbone~\cite{swin,pvt,lit,yang2021focal,cait} or adopting ViT as a strong module for modeling global relationships to address downstream tasks~\cite{segformer,vitdet,segmenter}. Another line of works focus on improving ViT efficiency by token pruning~\cite{hvt,dynamicvit}, quantization~\cite{fqvit,qvit} and dynamic inference~\cite{dynamicvit,WangHSHH21}, \etc. Most recently, large-scale self-supervised pretraining has helped ViTs achieve promising results on ImageNet, including contrastive learning~\cite{dino,mocov3} and masked image modeling~\cite{beit,mae,ibot,greenmim}. However, these models are designed to be over-parameterized and have a fixed computational cost, which is inflexible at the inference stage and cannot adapt to diverse and dynamic deployment environment. Instead of proposing or pretraining a new ViT architecture, we utilize different pretrained ViTs or even CNNs~\cite{resnet,resnext} to show that the proposed SN-Net is a general framework to assemble the existing model families.

\begin{figure*}[]
	\centering
	\includegraphics[width=1.0\linewidth]{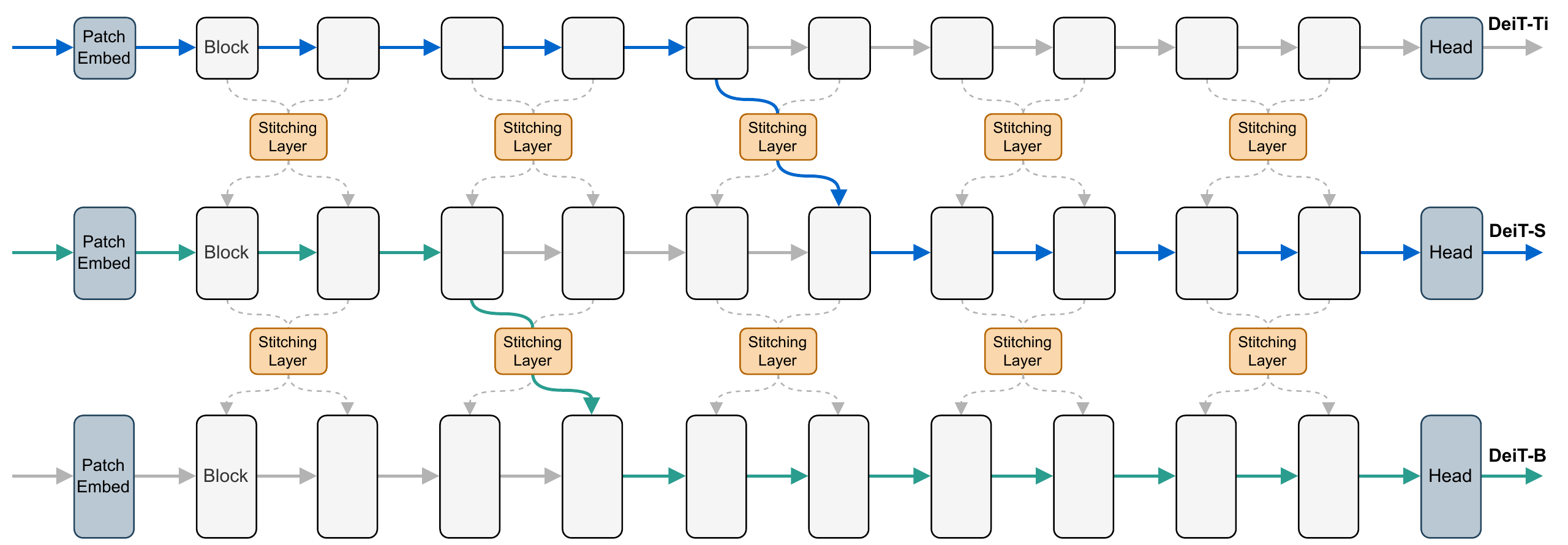}
	\caption{Illustration of the proposed \textbf{Stitchable Neural Network}, where three pretrained variants of DeiTs are connected with simple stitching layers ($1\times1$ convolutions). We share the same stitching layer among neighboring blocks (\eg, 2 blocks with a stride of 2 in this example) between two models. Apart from the basic anchor models, we obtain many sub-networks (stitches) by stitching the nearest pairs of anchors in complexity, \eg, DeiT-Ti and DeiT-S (the blue line), DeiT-S and DeiT-B (the green line). Best viewed in color.}
	\vspace{-10pt}
	\label{fig:framework}
\end{figure*}

\section{Method} \label{sec:method}
In this section, we first introduce the preliminary of model stitching at Section~\ref{sec:preliminary}. Next, we describe the details of our proposed stitchable neural networks at Section~\ref{sec:dsnnet}.

\subsection{Preliminaries of Model Stitching} \label{sec:preliminary}
Let $\theta$ be the model parameters of a pretrained neural network and $f_i$ represent the function of the $i$-th layer. A typical feed-forward neural network with $L$ layers can be defined as a composition of functions: $f_\theta = f_L \circ \cdot\cdot\cdot \circ f_1 $, where $\circ$ indicates the composition,  and $f_\theta : \mathcal{X} \to \mathcal{Y}$ maps the inputs in an input space  $\mathcal{X}$ to the output space $\mathcal{Y}$. Let $\mathbf{X} \in \mathcal{X}$ be an input to the network. The basic idea of model stitching involves splitting a neural network into two portions of functions at a layer index $l$. The first portion of layers compose the front part that maps the input $\mathbf{X}$ into the activation space of the $l$-th layer $ \mathcal{A}_{\theta, l}$, which can be formulated as
\begin{equation} \label{eq:front_func}
    H_{\theta,l}(\mathbf{X}) = f_l \circ \cdot\cdot\cdot \circ f_1 (\mathbf{X}) =  \mathbf{X}_l, 
\end{equation}
where $\mathbf{X}_l \in \mathcal{A}_{\theta, l}$ denotes the output feature map at the $l$-th layer. Next, the last portion of layers maps $\mathbf{X}_l$ into the final output
\begin{equation} \label{eq:tail_func}
    T_{\theta,l}(\mathbf{X}_l) = f_L \circ \cdot\cdot\cdot \circ f_{l+1} (\mathbf{X}_l). 
\end{equation}
In this case, the original neural network function $f_\theta$ can be defined as a composition of the above functions $f_\theta =  T_{\theta,l} \circ H_{\theta,l}$ for all layer indexes $l = 1, ..., L-1$. 

Now suppose we have another pretrained neural network $f_\phi$. Let $\mathcal{S} : \mathcal{A}_{\theta, l} \to \mathcal{A}_{\phi,m}$ be a stitching layer which implements a transformation between the activation space of the $l$-th layer of $f_\theta$ to the activation space of the $m$-th layer of $f_\phi$. The basic idea of model stitching to obtain a new network
defined by $\mathcal{S}$ can be expressed as
\begin{equation} \label{eq:stitch}
    F_S(\mathbf{X}) = T_{\phi,m} \circ \mathcal{S} \circ H_{\theta,l}(\mathbf{X}).
\end{equation}
By controlling the stitched layer indexes $l$ and $m$, model stitching can produce a sequence of stitched networks.
It has been observed by~\cite{stitch_0} that models of the same architecture but with different initializations (\ie, random seeds) can be stitched together with low loss penalty.
Further experiments by \cite{stitch_1,stitch_2} have demonstrated that different architectures (\eg, ViTs and CNNs) may also be stitched without significant performance drop, regardless they are trained in different ways such as self-supervised learning or supervised learning.

\subsection{Stitchable Neural Networks} \label{sec:dsnnet}
Based on the insight of model stitching, we propose Stitchable Neural Networks (SN-Net), a new ``many-to-many'' elastic model paradigm. SN-Net is motivated by an increasing number of pretrained models in the publicly available model zoo~\cite{rw2019timm}, where most of the individually trained models are not directly adjustable to dynamic resource constraints. To this end, SN-Net inserts a few stitching layers to smoothly connect a family of pretrained models to form diverse stitched networks permitting run-time network selection.
The framework of SN-Net is illustrated in Figure~\ref{fig:framework} by taking plain ViTs~\cite{vit} as an example. For brevity, we will refer to the models that to be stitched as ``\textbf{anchors}'' and the derived models by stitching anchors  as ``\textbf{stitches}''.
In the following, we describe the concrete approach in detail, including what, how and where to stitch, the stitching strategy and space, as well as an effective and efficient training strategy for SN-Net.

\paragraph{What to stitch: the choice of anchors.}
In general, the large-scale model zoo determines the powerful representation capability of SN-Net as it is a universal framework for assembling the prevalent families of architectures.
As shown in Section~\ref{sec:experiment}, SN-Net works for stitching representative ViTs and CNNs.
However, intuitively, anchors that are pretrained on different tasks can learn very different representations (\eg, ImageNet~\cite{imagenet} and COCO~\cite{coco}) due to the large distribution gap of different domains~\cite{transfer_PanY10}, thus making it difficult for stitching layers to learn to transform activations among anchors. Therefore, the selected anchors should be consistent in terms of the pretrained domain.

\paragraph{How to stitch: the stitching layer and its initialization.}
Conceptually, the stitching layer should be as simple as possible since its aim is not to improve the model performance, but to transform the feature maps from one activation space to another~\cite{stitch_1}. To this end, the stitching layers in SN-Net are simply $1\times1$ convolutional layers.
By default in PyTorch~\cite{pytorch}, these layers are initialized based on Kaiming initialization~\cite{he_init}. 

However, different from training a network from scratch as in most works~\cite{swin,pvt,beit,cait}, SN-Net is built upon pretrained models. In this case, the anchors have already learned good representations, which allows to directly obtain an accurate transformation matrix by solving the following least squares problem
\begin{equation} \label{eq:least_square}
    \| \mathbf{A}\mathbf{M}_o-\mathbf{B}\|_F= \min\| \mathbf{A}\mathbf{M}-\mathbf{B}\|_F,
\end{equation}
where $\mathbf{A} \in  \mathbb{R}^{N \times D_1}$ and $\mathbf{B} \in \mathbb{R}^{N \times D_2}$ are two feature maps of the same spatial size but with different number of hidden dimensions. $N$ denotes the length of the input sequence and $D_1, D_2$ refer to the number of hidden dimensions. $\mathbf{M} \in  \mathbb{R}^{D_1 \times D_2}$ is the targeted transformation matrix.

One can observe that Eq.~(\ref{eq:least_square}) indicates a closed form expression based on singular value decomposition, in which case the optimal solution can be achieved through an orthogonal projection in the space of matrices,
\begin{equation} \label{eq:mp_inverse}
    \mathbf{M}_o = \mathbf{A}^{\dagger}\mathbf{B},
\end{equation}
where $\mathbf{A}^{\dagger}$ denotes the Moore-Penrose pseudoinverse of $\mathbf{A}$. To obtain $\mathbf{M}_o$ requires only a few seconds on one CPU with hundreds of samples. 
However, we will show in Section~\ref{sec:ablation} that directly using with least-squares solution achieves unstable performance for stitches, but it actually provides a good initialization for learning stitching layers with SGD. Therefore, the least-squares solution serves as the default initialization approach for the stitching layers in SN-Net.

\begin{figure}[]
	\centering
	\includegraphics[width=0.85\linewidth]{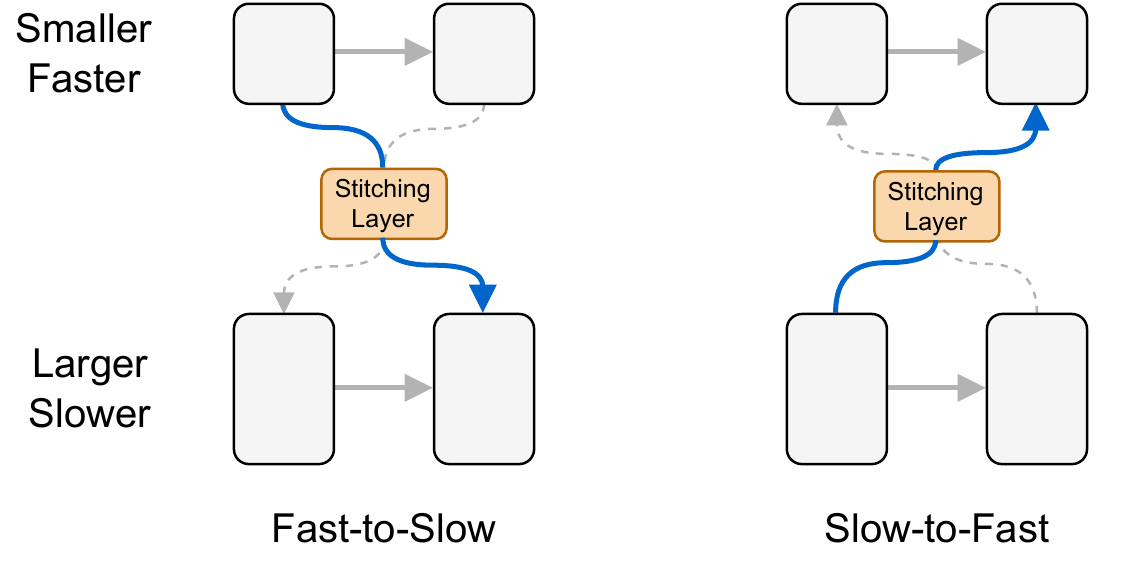}
	\caption{Stitching direction: Fast-to-Slow \vs Slow-to-Fast.}
	\vspace{-15pt}
	\label{fig:stitch_direction}
\end{figure}

\paragraph{Where to stitch: the stitching directions.}
Given anchors with different scales and complexities, 
there are two options to stitch them together: \textbf{Fast-to-Slow} and \textbf{Slow-to-Fast}. Taking two anchors as an example (Figure~\ref{fig:stitch_direction}), Fast-to-Slow takes the first portion of layers (\ie, Eq.~(\ref{eq:front_func})) from a smaller and faster model, and the last portion of layers (\ie, Eq.~(\ref{eq:front_func}))  from a larger and slower model, where Slow-to-Fast goes in a reverse direction. 
However, as Fast-to-Slow is more aligned with the existing model design principle (\ie, increasing the network width as it goes deeper), we will show in Section~\ref{sec:ablation} that it achieves more stable and better performance than Slow-to-Fast. In this case, we take Fast-to-Slow as the default stitching direction in SN-Net. Besides, as different anchors may reach very different minima, we propose a \textbf{nearest stitching} strategy by limiting the stitching between two anchors of the nearest model complexity/performance. Thus, each stitch in SN-Net assembles a pair of neighbouring anchors. We will show in Section~\ref{sec:ablation} that stitching across anchors without the nearest stitching constraint achieves inferior performance.

\begin{figure}[]
	\centering
	\includegraphics[width=1.0\linewidth]{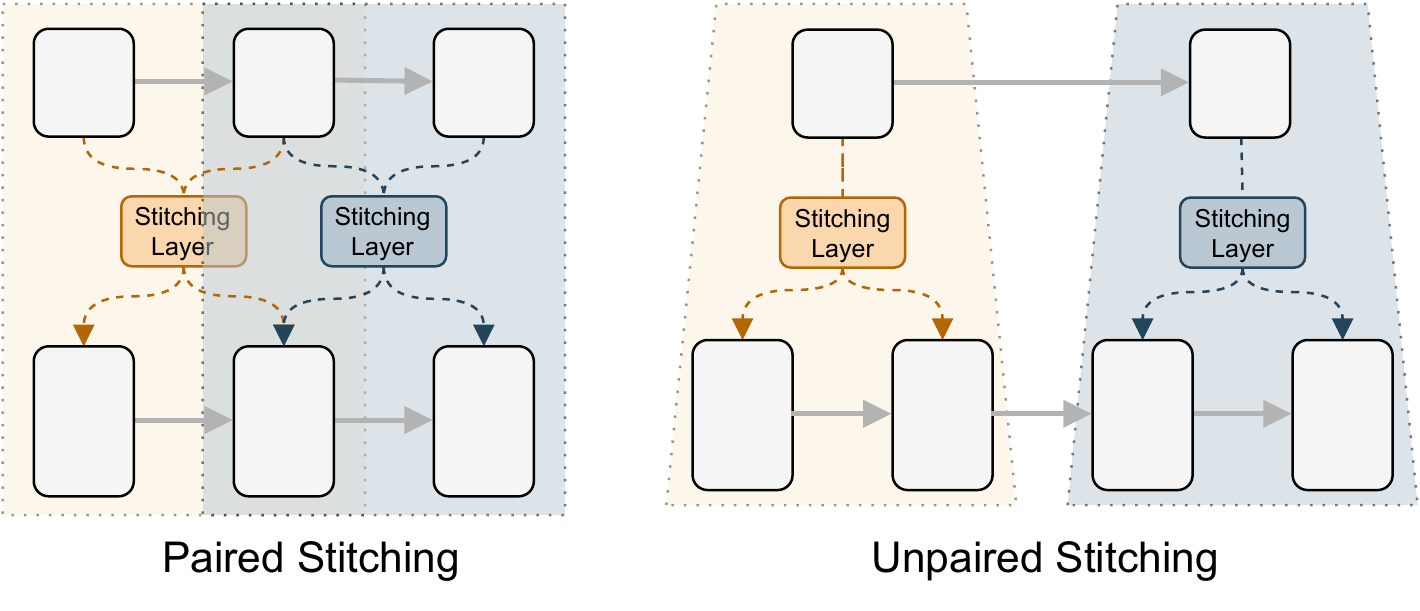}
	\vspace{-15pt}
	\caption{Stitching as sliding windows, where paired stitching is proposed for stitching models with equal depth and unpaired stitching is utilised for models with unequal depth.}
	\vspace{-15pt}
	\label{fig:stitch_strategy}
\end{figure}

\paragraph{Way to stitch: stitching as sliding windows.} Our stitching strategy is inspired by the main observation: neighboring layers dealing with the same scale feature maps share similar representations~\cite{cka}. To this end, we propose to stitch anchors as sliding windows, where the same window shares a common stitching layer, as shown in Figure~\ref{fig:stitch_strategy}. Let $L_1$ and $L_2$ be depth of two anchors. Then intuitively, there are two cases when stitching layers/blocks between the two anchors: \textbf{paired stitching} ($L_1 = L_2$) and \textbf{unpaired stitching} ($L_1 \neq  L_2$). In the case of $L_1 = L_2$, the sliding windows can be controlled as sliding windows with a window size $k$ and a stride $s$. Figure~\ref{fig:stitch_strategy} left shows an example with $k=2, s=1$. However, in most cases we have unequal depth as different model architectures have different scales. Even though, matching $L_1$ layers to $L_2$ layers can be easily done by nearest interpolation where each layer from the shallower anchor can be stitched with more than one layers of the deeper anchor, as shown in Figure~\ref{fig:stitch_strategy} right.  

\paragraph{Stitching space.}
In SN-Net, we first split the anchors along the internal layers/blocks at each stage then apply our stitching strategy within each stage. As different anchors have different architectural configurations, the size of the stitching space can be variant based on the depth of the selected anchors and the stitching settings (\ie, the kernel size $k$ and stride $s$). For example, with $k=2$ and $s=1$, DeiT-based SN-Net can have 71 stitches under the constraint of our nearest stitching principle, or 732 stitches without this constraint, as shown in Figure~\ref{fig:compare_three_ablate} (b). We provide detailed illustrations for this figure in the supplementary material. More stitches can be obtained by choosing anchors with larger scales or configuring the sliding windows by using a larger window size or smaller stride.
Overall, compared to one-shot NAS which can support more than $10^{20}$ sub-networks, SN-Net has a relatively smaller space (up to hundreds or thousands). However, we point out that even though NAS has a much larger architecture space, during deployment, it only focuses on the sub-networks on the Pareto frontier of performance and resource consumption~\cite{bignas}. Thus the vast majority of sub-networks are ignored. 
In contrast, we will show in Section~\ref{sec:ablation} that the stitches in SN-Net smoothly distribute among the anchors, which indicates that the analogous performance curve can almost be estimated without too much searching cost, permitting fast deployment.

\paragraph{Training strategy.}
Given the anchors with different accuracy-efficiency trade-offs from the model zoo, our aim is to train an elastic joint network that covers a large number of stitches in a highly efficient way so that it can fit diverse resource constraints with low energy cost. 
The detailed training algorithm is provided in Algorithm~\ref{algos:training} with PyTorch style, where we firstly define a configuration set that contains all possible stitches and initialize all stitching layers with least-squares matching by solving Eq.~(\ref{eq:least_square}). Next, at each training iteration, we randomly sample a stitch and follow the standard training process as in common practices~\cite{swin,pvt}. To further improve the performance of stitches, we also adopt knowledge distillation with RegNetY-160~\cite{regnet} as the teacher model. 
The overall training process requires only a few epochs (\eg, 50) on ImageNet, which is far less than the supernet training in NAS~\cite{CaiZH19,XieZLL19,ofa} and other techniques~\cite{slimmable_nn_v1,slimmable_v2} that train networks from scratch. Moreover, as anchors are already well-trained, we do not observe significant interference~\cite{ofa} among stitches, as shown in the experiments.

\begin{algorithm}[ht]
\caption{Training Stitchable Neural Networks}
    \begin{algorithmic}[1]
    \Require{$M$ pretrained anchors to be stitched. Configuration set $E = \{e_1, ..., e_Q\}$ with $Q$ stitching positions.}
    \State{Initialize all stitching layers by least-squares matching}
    \For {$i = 1, ..., n_{iters}$}
        \State{Get next mini-batch of data $\mathbf{X}$ and label $\mathbf{Y}$.}
        \State{Clear gradients, \(optimizer.zero\_grad()\).}
        \State{Randomly sample a stitching $e_q$ from set $E$.}
        \State{Execute the current stitch, \(\hat{\mathbf{Y}} = F_{e_{q}}(\mathbf{X})\).}
        \State{Compute loss, \(loss = criterion(\hat{\mathbf{Y}}, \mathbf{Y})\).}
        \State{Compute gradients, \(loss.backward()\).}
        \State{Update weights, \(optimizer.step()\).}
    \EndFor
    \end{algorithmic}
\label{algos:training}
\end{algorithm}

\begin{table*}[]
\centering
\caption{Performance comparisons on ImageNet-1K between individually trained models from scratch with \textbf{300 epochs} and stitches selected from our proposed SN-Net trained with \textbf{50 epochs}. A single SN-Net with 118.4M parameters can include all possible stitches. We denote ``\# Ti/S/B Blocks'' as the number of stitched blocks chosen from DeiT-Ti/S/B, respectively. ``failed'' means training such stitched model from scratch fails to converge and incurs ``loss is nan''. Throughput is measured on one RTX 3090 and averaged over 30 runs, with a batch size of 64 and input resolution of $224\times224$.}
\vspace{-10pt}
\label{tab:main_deit_res}
\scalebox{0.95}{
\begin{tabular}{ccc|cc|cc|cc}
\multirow{2}{*}{\textbf{\# Ti Blocks}} & \multirow{2}{*}{\textbf{\# S Blocks}} & \multirow{2}{*}{\textbf{\# B Blocks}} & \multirow{2}{*}{\textbf{\begin{tabular}[c]{@{}c@{}}FLOPs \\ (G)\end{tabular}}} & \multirow{2}{*}{\textbf{\begin{tabular}[c]{@{}c@{}}Throughput \\ (images/s)\end{tabular}}} & \multicolumn{2}{c|}{\textbf{Individually Trained}} & \multicolumn{2}{c}{\textbf{SN-Net}} \\ \cline{6-9} 
 &  &  &  &  & \textbf{Params (M)} & \textbf{Top-1 (\%)} & \textbf{Params (M)} & \textbf{Top-1 (\%)} \\ \shline
12 & 0 & 0 & 1.3 & 2,839 & 5.7 & 72.1 & \multirow{9}{*}{118.4} & 70.6 \\
9 & 3 & 0 & 2.1 & 2,352 & 10.0 & 75.9 &  & 72.6 \\
6 & 6 & 0 & 2.9 & 1,963 & 14.0 & 78.2 &  & 76.5 \\
3 & 9 & 0 & 3.8 & 1,673 & 18.0 & 79.4 &  & 78.2 \\
0 & 12 & 0 & 4.6 & 1,458 & 22.1 & 79.8 &  & 79.5 \\
0 & 9 & 3 & 7.9 & 1,060 & 38.7 & 79.4 &  & 80.0 \\
0 & 6 & 6 & 11.2 & 828 & 54.6 & failed &  & 81.5 \\
0 & 3 & 9 & 14.3 & 679 & 70.6 & 80.3 &  & 82.0 \\
0 & 0 & 12 & 17.6 & 577 & 86.6 & 81.8 &  & 81.9
\end{tabular}
}
\vspace{-5pt}
\end{table*}

\vspace{-10pt}
\section{Experiment}
\label{sec:experiment}

\paragraph{Implementation details.}
We conduct all experiments on ImageNet-1K~\cite{imagenet}, a large-scale image dataset which contains $\sim$1.2M training images and 50K validation images from 1K categories. Model performance is measured by Top-1 accuracy. Furthermore, we report the FLOPs and throughput as indicators of theoretical complexity and real speed on hardware, respectively. We study stitching plain ViTs, hierarchical ViTs, CNNs, and CNN with ViT. We choose the representative model families as anchors: DeiT~\cite{deit}, Swin Transformer~\cite{swin} and ResNet~\cite{resnet}. By default, we randomly sample 100 images from the training set to initialize the stitching layers. For paired stitching, we set the default sliding kernel size as 2 and the stride as 1. For unpaired stitching, we match layers by nearest interpolation. Unless otherwise specified, all experiments adopt a total batch size of 1,024 on 8 V100 GPUs. 
We train DeiT/Swin with 50 epochs with an initial learning rate of $1\times10^{-4}$. For the experiments with ResNet, we train with 30 epochs based on the training scripts from timm~\cite{rw2019timm} with an initial learning rate of $0.05$. All other hyperparameters adopt the default setting as in~\cite{deit,swin,rw2019timm}. For hierarchical models, we scale the learning rate of the anchor parameters by $1/10$ compared to that of stitching layers.

\subsection{Main Results} \label{sec:main_exp}
\begin{figure*}[]
	\centering
	\includegraphics[width=\linewidth]{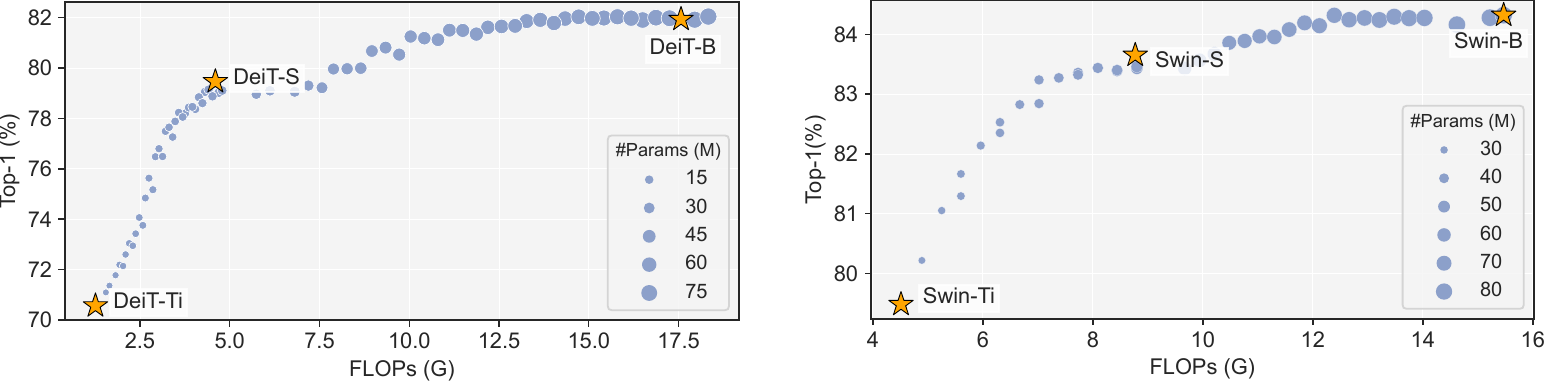}
	\vspace{-15pt}
	\caption{Performance of SN-Net by stitching DeiT-Ti/S/B and Swin-Ti/S/B.}
	\vspace{-10pt}
	\label{fig:main_deit_swin_res}
\end{figure*}

\paragraph{Stitching plain ViTs.}
Based on Algorithm~\ref{algos:training}, we first generate a stitching configuration set by assembling ImageNet-1K pretrained DeiT-Ti/S/B, which contains 71 stitches including 3 anchors. Then we jointly train the stitches in DeiT-based SN-Net on ImageNet with 50 epochs. The whole training and evaluation process takes only around 110 and 3 GPU hours on V100 GPUs, respectively. In Figure~\ref{fig:main_deit_swin_res} left, we visualize the performance of all 71 stitches, including the anchors DeiT-Ti/S/B (highlighted as yellow stars). In general, SN-Net achieves a wide range of successful stitches, where they achieve smoothly increased performance when stitching more blocks from a larger anchor. We also observe a phenomenon of \textbf{model-level interpolation} between two anchors: \textit{with the architectures of the stitches become more similar to the nearest larger anchor, the performance also gradually gets closer to it.}

Moreover, we compare individually trained models from scratch and select stitches from our jointly optimized SN-Net. For brevity, we denote ``Ti-S'' as the stitches with DeiT-Ti/S as anchors and ``S-B'' as the stitches with DeiT-S/B as anchors. The results are shown in Table~\ref{tab:main_deit_res}. As it indicates, compared to individually trained ``S-B'' stitches, SN-Net achieves even better performance. It is worth noting that some stitches can fail to converge when training from scratch. However, due to all anchors in SN-Net have been well-trained, the stitches can be easily interpolated among them. Also note that ``Ti-S'' stitches achieve inferior performance than individual ones. We speculate that due to a slightly larger performance gap between DeiT-Ti/S compared to DeiT-S/B, training Ti-S stitches from scratch may help to find a better local optimum. We also notice a performance drop for anchor DeiT-Ti, for which we assume a more intelligent stitch sampling strategy can help in future works. Overall, a single SN-Net can cover a wide range of accuracy-efficiency trade-offs while achieving competitive performance with models that trained from scratch. To be emphasized, SN-Net reduces around $22\times$ training cost ($71 \times 300$ epochs \vs $3 \times 300 + 50$ epochs) and local disk storage (2,630M \vs 118M) compared to training and saving all individual networks.

\noindent\textbf{Stitching hierarchical ViTs.}
Furthermore, we conduct experiment by stitching hierarchical ViTs. In particular, we assemble Swin-Ti/S/B trained on ImageNet-22K  by stitching the blocks at the first three stages.  Note that we do not choose ImageNet-1K pretrained Swin models due to the minor performance gap (83.1\% \vs 83.5\%) but the large difference in FLOPs (8.7G \vs 15.4G) between Swin-S/B. We visualize the results at Figure~\ref{fig:main_deit_swin_res} right. It shows that the Swin-based SN-Net also achieves flexible accuracy-efficiency trade-offs among the three anchors. This strongly demonstrates that the proposed SN-Net is a general solution for both plain and hierarchical models.

\begin{figure}[]
	\centering
	\includegraphics[width=\linewidth]{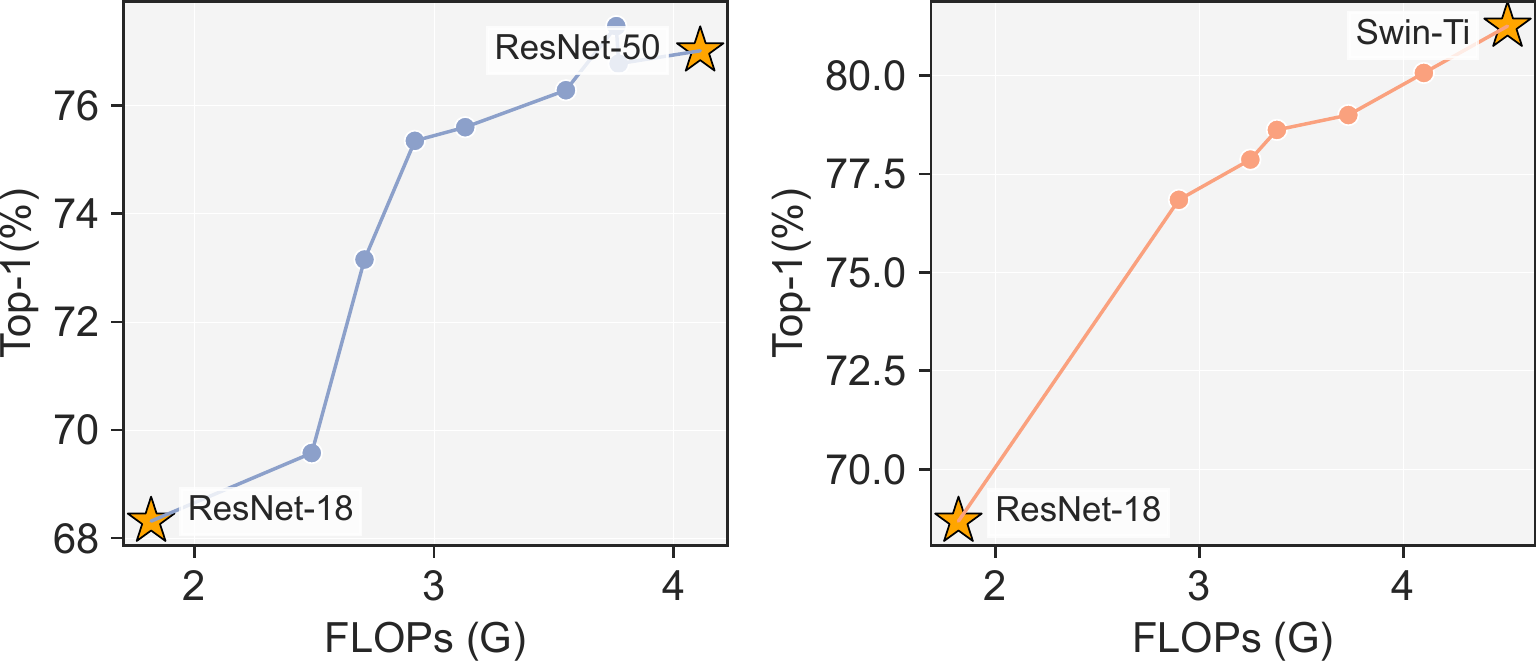}
	\caption{Effect of stitching CNNs and CNN-ViT.}
	\vspace{-15pt}
	\label{fig:cnn_vit}
\end{figure}

\noindent\textbf{Stitching CNNs and CNN-ViT.}
We show that SN-Net also works for stitching CNN models and even connecting CNNs with ViTs. As Figure~\ref{fig:cnn_vit} shows, with only 30 epochs of training, the stitches by assembling from ResNet-18~\cite{resnet} to ResNet-50/Swin-Ti perform favourably, which again emphasizes that SN-Net can be general for both CNNs and ViTs. Also note that ResNet-18/50 and Swin-Ti are shallow models, so we obtain a small number of stitches.

\begin{figure}[]
	\centering
	\includegraphics[width=\linewidth]{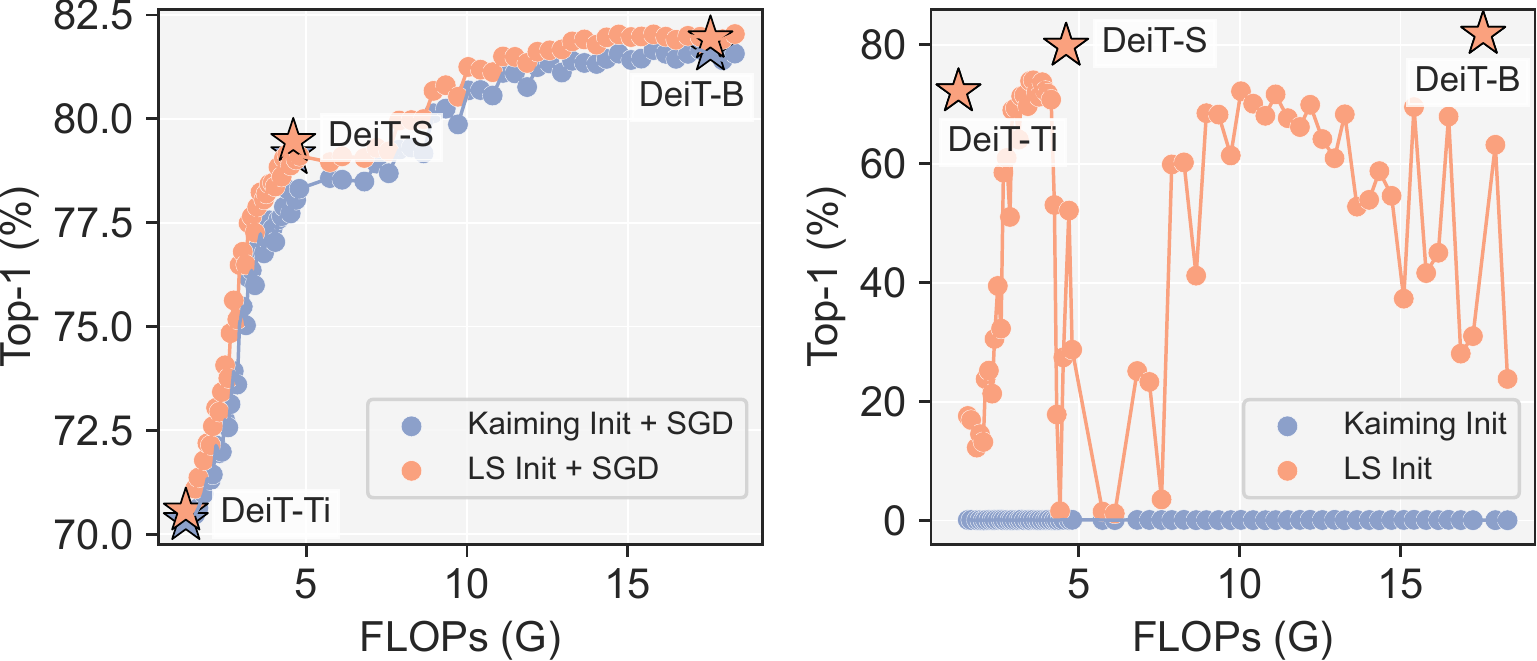}
	\caption{Different learning strategies for stitching layers.}
	\vspace{-15pt}
	\label{fig:compare_stitching_learning}
\end{figure}

\begin{figure*}[]
	\centering
	\includegraphics[width=\linewidth]{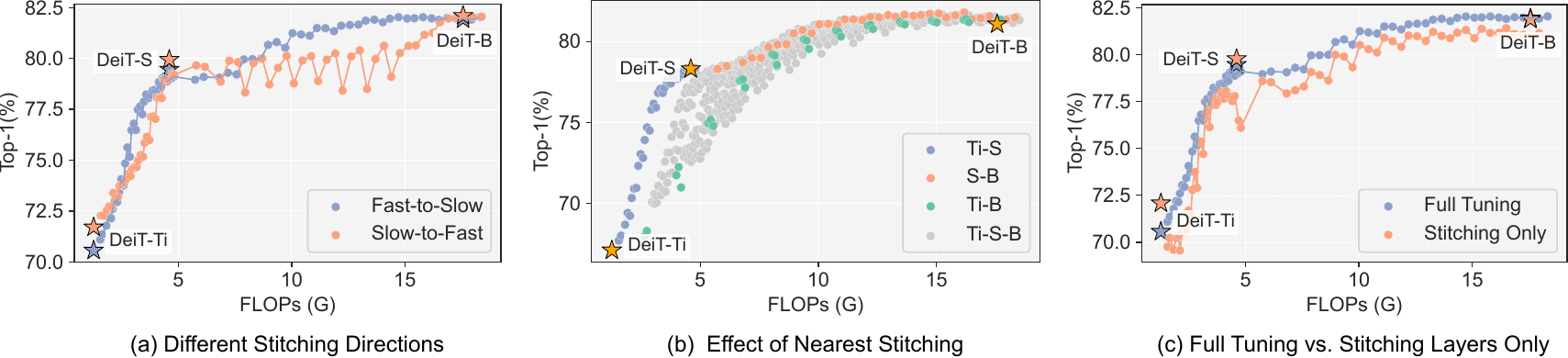}
	\vspace{-15pt}
	\caption{From left to right, Figure (a) shows the effect of different stitching directions. Figure (b) presents the effect of nearest stitching based on DeiT, where ``Ti'', ``S'', ``B'' denote the stitched anchors. For example, ``Ti-S-B'' refers to a stitch that defined by connecting the tiny, small and base variants of DeiT, sequentially. Figure (c) shows the comparison of full model tuning vs. tuning stitching layers only. }
	\vspace{-15pt}
	\label{fig:compare_three_ablate}
\end{figure*}

\subsection{Ablation Study} \label{sec:ablation}
In this section, we ablate the design principles for SN-Net. Unless otherwise specified, our experiments are based on stitching DeiT-Ti/S/B with $k=2, s=1$ and knowledge distillation with RegNetY-160. By default, the training strategy is the same as in Section~\ref{sec:main_exp}, \eg, 50 epochs on ImageNet. We provide more ablation studies in the supplementary material, such as the effect of kernel size and stride for controlling the sliding windows during stitching, \etc.

\noindent\textbf{Effect of different stitching layer learning strategies.}
To study the effect of different learning strategies for stitching layers, we consider 4 cases: 1) \textbf{Kaiming Init}, the default initialization method in PyTorch. 2) \textbf{Least-squares (LS) Init}, the LS solution by solving Eq.~(\ref{eq:least_square}). 3) \textbf{Kaiming Init + SGD}, learning with gradients update on ImageNet after Kaiming Init. 4) \textbf{LS Init + SGD}, learning with gradients update on ImageNet after LS Init. We report the experiment results in Figure~\ref{fig:compare_stitching_learning}. Overall, we find LS Init serves as a good starting point for learning stitching layers compared to the default Kaiming Init. Interestingly, we observe some stitches by directly matching with LS solution perform quite well compared to Kaiming Init, as shown in Figure~\ref{fig:compare_stitching_learning} right. However, in general, directly matching with LS solution results in an unstable performance curve. This indicates that LS Init is not fully aware of the final performance of the stitches and updating the stitching layers is essential.

\noindent\textbf{Effect of stitching directions.}
In Figure~\ref{fig:compare_three_ablate} (a), we compare the stitching directions of Fast-to-Slow and Slow-to-Fast based on DeiT. In general, Fast-to-Slow helps to ensure a better performance for most stitches. On the other hand, Slow-to-Fast obtains a more unstable performance curve, especially for stitching DeiT-S/B. Compared to Fast-to-Slow which increases feature representation capacity by expanding the hidden dimension of activations from a narrower model to a wider one, Slow-to-Fast shrinks the hidden dimension, which contradicts to the existing model design principle~\cite{regnet,resnet,densenet} that gradually expands the hidden dimension to encode rich semantics as the network goes deeper. Therefore, the resulting information loss of Slow-to-Fast may increase the optimization difficulty.

\noindent\textbf{Effect of nearest stitching.}
In SN-Net, we adopt the nearest stitching strategy which limits a stitch to connect with a pair of anchors that have the nearest model complexity/performance. However, it is possible to simultaneously stitching more than two anchors (\eg, stitching all DeiT-Ti/S/B sequentially) or anchors with a large gap in complexity/performance (\eg, stitching DeiT-Ti with DeiT-B). With the same 50 epochs training, this approach helps to produce 10$\times$ more stitches than our default settings (732 \vs 71). However, as shown in Figure~\ref{fig:compare_three_ablate} (b), even though Ti-B and Ti-S-B achieve good interpolated performance among the anchors (\ie, they are stitchable), most of them cannot outperform Ti-S and S-B stitches. In the case of Ti-B, we speculate that without a better minima as a guide in the middle (\eg, DeiT-S), the local minima that found by stitching layers can be sub-optimal due to the large complexity/performance gap between two anchors. Besides, stitching more than two anchors simultaneously does not bring obvious gain at this stage, which we leave for future work.

\noindent\textbf{Effect of tuning full model vs. stitching layers only.}
In SN-Net, the role of stitching layers is to map the feature maps from one activation space to another. However, since the anchors have been well-trained, one question is how the performance changes if we only update the stitching layers during training. In Figure~\ref{fig:compare_three_ablate} (c), we show that tuning stitching layers is only promising for some stitches. In contrast, we observe that the performance of stitches can be improved by tuning the full model. Therefore, we by default make SN-Net to be fully updated during training.
\section{Conclusion}
\label{sec:conclusion}
We have introduced Stitchable Neural Networks, a novel general framework for developing elastic neural networks that directly inherit the rich knowledge from pretrained model families in the large-scale model zoo. Extensive experiments have shown that SN-Net can deliver fast and flexible accuracy-efficiency trade-offs at runtime with low cost, fostering the massive deployment of deep models for real-world applications.
With the rapid growth of the number of large-scale pretrained models~\cite{mae,clip}, we believe our work paves a new way for efficient model development and deployment, yielding a significant step towards Green AI. In future works, SN-Net can be extended into more tasks, such as natural language processing, dense prediction and transfer learning.

\noindent\textbf{Limitations and societal impact.} 
Our current training strategy randomly samples a stitch at each training iteration, which implies that with a much larger stitching space, the stitches may not be sufficiently trained unless using more training epochs. We leave this for future work.

\appendix
\section*{\centering{Appendix}}
We organize our supplementary material as follows. 
\begin{itemize}
    \item In Section~\ref{sec:supp_nearest_stitch}, we provide further explanation of the proposed nearest stitching strategy.
    \item In Section~\ref{sec:eff_sliding_window}, we study the effect of different sizes and strides of sliding windows for stitching.
    \item In Section~\ref{sec:diff_epochs}, we study the effect of different training epochs.
    \item In Section~\ref{sec:sandwich}, we show the effectiveness of our training strategy by comparing with sandwich sampling rule and inplace distillation~\cite{slimmable_v2}.
    \item In Section~\ref{sec:train_wo_prerained}, we discuss the effect of training without the pretrained weights of anchors.
    \item In Section~\ref{sec:eff_diff_sample_init}, we experiment with different number of samples for initializing stitching layers.
    \item In Section~\ref{sec:compare_nas}, we provide additional discussion with One-shot NAS.
    \item In Section~\ref{sec:compare_layerdrop}, we compare SN-Net with LayerDrop~\cite{layerdrop} at inference time.
\end{itemize}

\section{Detailed Illustration of Nearest Stitching Strategy} \label{sec:supp_nearest_stitch}
\begin{figure*}[!htb]
	\centering
	\includegraphics[width=1.0\linewidth]{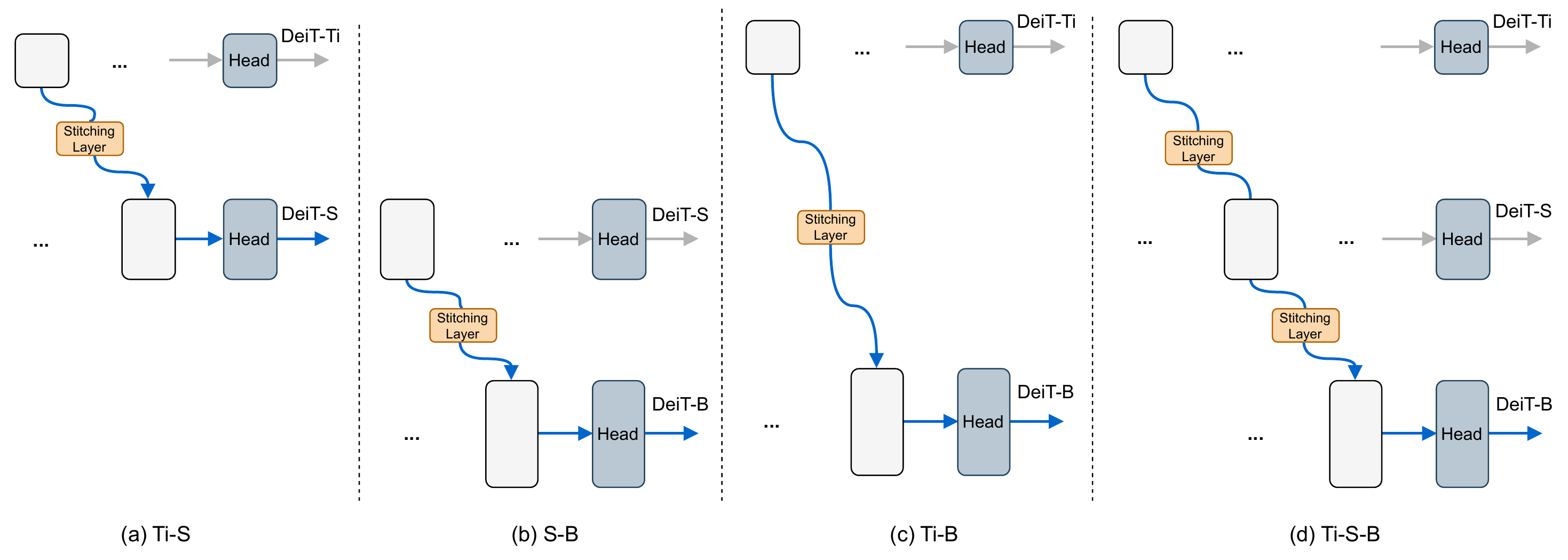}
	\caption{Four types of stitches based on DeiT-Ti/S/B.
	Under the proposed nearest stitching strategy, we limit the stitching between two anchors of the nearest model complexity/performance, \ie, Figure (a) and (b), while excluding stitching anchors with a larger complexity/performance gap (Figure (c)) or sequentially stitching more than two anchors (Figure (d)).}
	\label{fig:supple_nearest_stitch}
\end{figure*}

In the proposed SN-Net, we introduce a nearest stitching strategy which limits the stitching between two anchors of the nearest complexity/performance.
In Figure~\ref{fig:supple_nearest_stitch}, we describe more details for this approach based on DeiT~\cite{deit}. Under the nearest stitching, we limit the stitches to two types: Ti-S and S-B, which connects DeiT-Ti/S and DeiT-S/B, respectively. Experiments in the main manuscript have shown that stitching anchors with a larger complexity/performance gap or sequentially stitching more than two anchors achieves inferior performance.

\section{Effect of Different Sizes and Strides of Sliding Windows} \label{sec:eff_sliding_window}
\begin{figure*}[!htb]
	\centering
	\includegraphics[width=\linewidth]{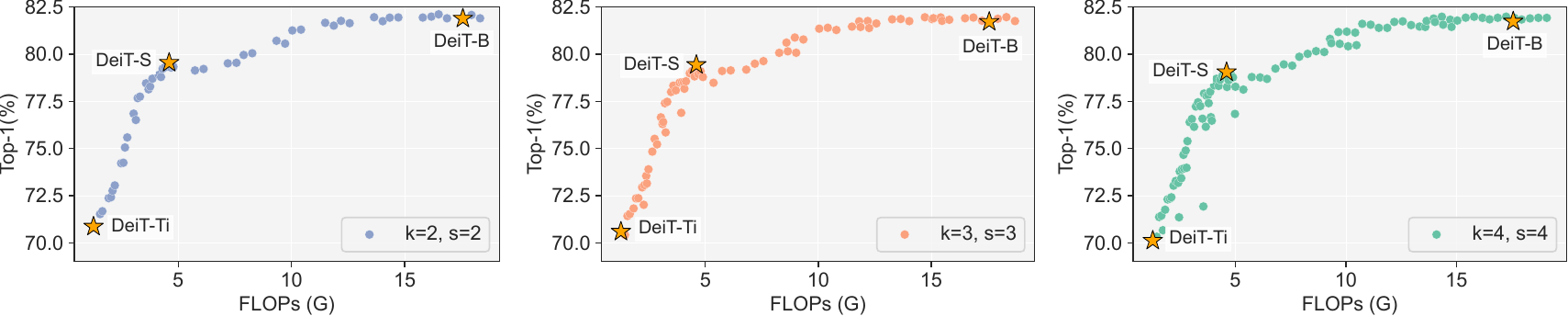}
	\caption{Effect of different sizes of sliding windows. $k$ and $s$ refer to the kernel size and stride for controlling the sliding windows. From left to right, the kernel sizes and strides of 2, 3 and 4 produce 51, 75 and 99 stitches, respectively.}
	\vspace{-10pt}
	\label{fig:compare_sliding_window}
\end{figure*}

We explore different settings of sliding windows in SN-Net. In Figure~\ref{fig:compare_sliding_window}, we visualize the results of using different kernel sizes and strides in stitching DeiT models. Overall, different settings can produce a different number of stitches but achieve similar good performance. However, it is worth noting that within a larger window, a stitching layer needs to map activations with more dissimilar representations, which potentially results in some bad-performed stitches, as shown in the case of $k=4, s=4$ in Figure~\ref{fig:compare_sliding_window}.

\vspace{-5pt}
\section{Effect of Different Training Epochs} \label{sec:diff_epochs}
\begin{figure}[!htb]
	\centering
	\includegraphics[width=\linewidth]{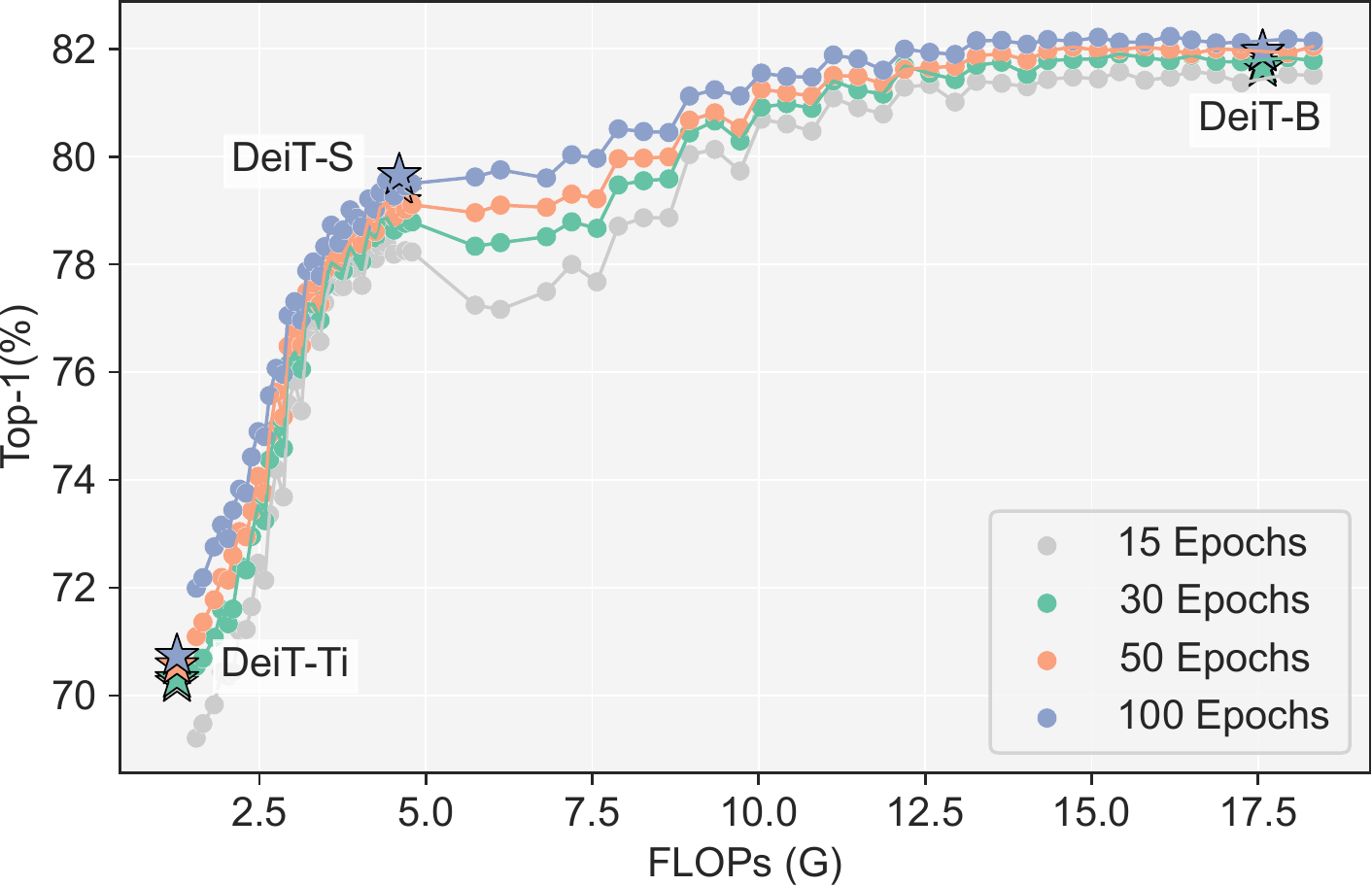}
	\caption{Effect of different training epochs.}
	\vspace{-10pt}
	\label{fig:compare_training_time}
\end{figure}

In our default setting, we train DeiT-based SN-Net with 50 epochs. In Figure~\ref{fig:compare_training_time}, we show that even with only 15 epochs, many stitches in SN-Net still perform favorably. With more training epochs, we observe consistent performance gain for all stitches, especially for those close to the smaller anchors, \eg. stitches around DeiT-Ti and DeiT-S. This is also reasonable as these stitches perform badly at the very beginning, thus particularly requiring more training time to obtain good performance.

\begin{figure}[!htb]
	\centering
	\includegraphics[width=\linewidth]{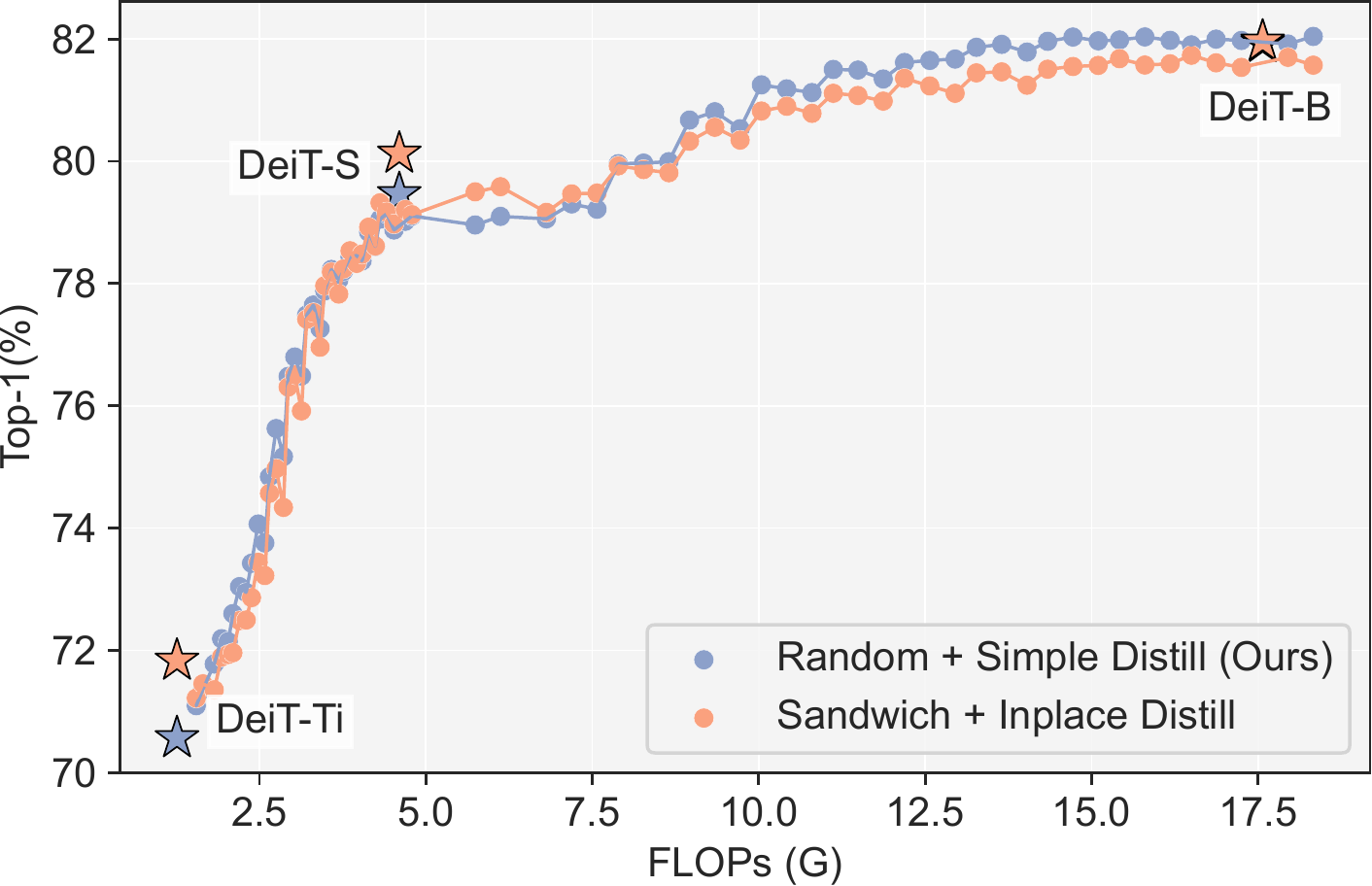}
	\caption{Comparison between our training strategy and common supernet training strategy in NAS (\ie, sandwich sampling rule and inplace distillation~\cite{slimmable_v2}).
	}
	\label{fig:compare_sandwich}
	\vspace{-10pt}
\end{figure}

\vspace{-5pt}
\section{Effect of Training Strategy} \label{sec:sandwich}
To train SN-Net, we adopt a simple training strategy by randomly sampling a stitch at each training iteration and using simple distillation for all stitches with a typical teacher (\eg, RegNetY-160~\cite{regnet}). To show the effectiveness of this strategy, we conduct experiments by training SN-Net with sandwich sampling rule and inplace distillation~\cite{slimmable_v2}, which is a common practice for training supernets in NAS~\cite{bignas,dy_slimmable,DBLP:conf/bmvc/PiWLLY21}. Specifically, we simultaneously sample one stitch and its connected pair of anchors at each training iteration. At the same time, we use the larger anchor as the teacher to guide the smaller anchor and the sampled stitch. However, as shown in Figure~\ref{fig:compare_sandwich}, this approach mainly improves the smaller anchors (\ie, DeiT-Ti/S) while most stitches cannot outperform those under our training strategy. It is also worth noting that the sandwich rule requires intensive memory/time cost due to training multiple networks at one training iteration. In contrast, ours requires a similar training cost for each training iteration as a normal network training~\cite{deit}.

\vspace{-5pt}
\section{Training without Pretrained Weights} \label{sec:train_wo_prerained}
The foundation of SN-Net is based on the pretrained model families in the large-scale model zoo. Without the pretrained weights of anchors, we find SN-Net failed to converge (training failed within 10 epochs based on the default experiment settings of DeiT), which aligns with our assumption that pretrained anchors help to reduce many training difficulties, such as the interference among stitches.

\begin{figure}[!htb]
	\centering
	\includegraphics[width=\linewidth]{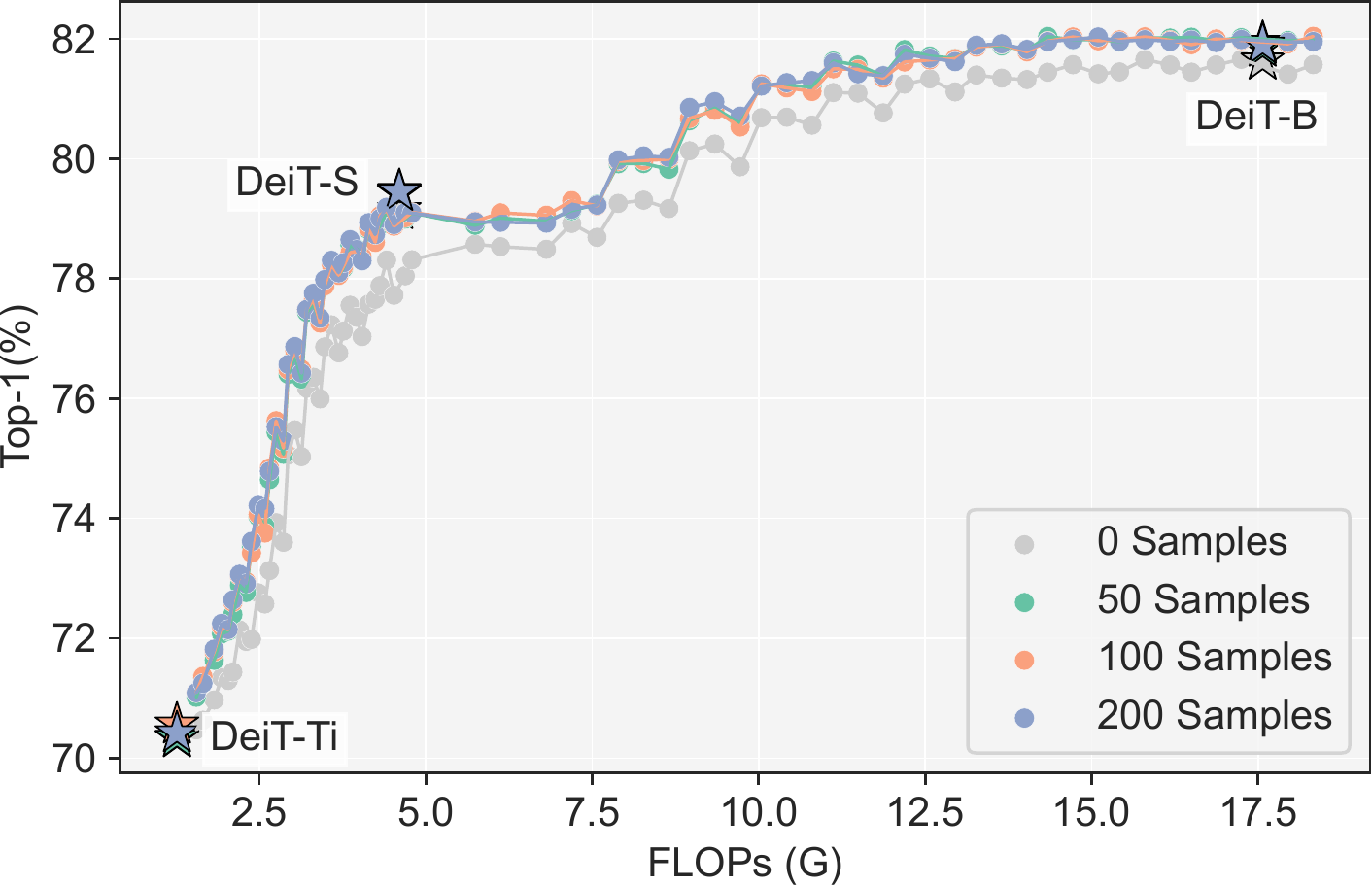}
	\caption{Effect of different number of samples for initializing stitching layers. With 0 samples, the initialization is equivalent to the default Kaiming initialization in PyTorch.}
	\label{fig:compare_init_samples}
\end{figure}

\section{Effect of Different Number of Samples for Initializing Stitching Layers} \label{sec:eff_diff_sample_init}
By default, we randomly sample 100 training images to initialize the stitching layers in SN-Net. To explore the effect of different number of samples for initialization, we train DeiT-based SN-Net by using 50, 100 and 200 training images on ImageNet with the same 50 epochs of training. As shown in Figure~\ref{fig:compare_init_samples}, although all settings achieve better performance than the default Kaiming initialization in PyTorch, we find that using more samples does not bring more performance gain. Besides, since solving the least-squares solution with more samples can increase the memory cost at the beginning of training, we set the default number of samples for initializing stitching layers to 100 to avoid the potential ``out of memory'' issue.

\section{Compared with One-shot NAS} \label{sec:compare_nas}
\begin{figure}[htb!]
	\centering
	\includegraphics[width=\linewidth]{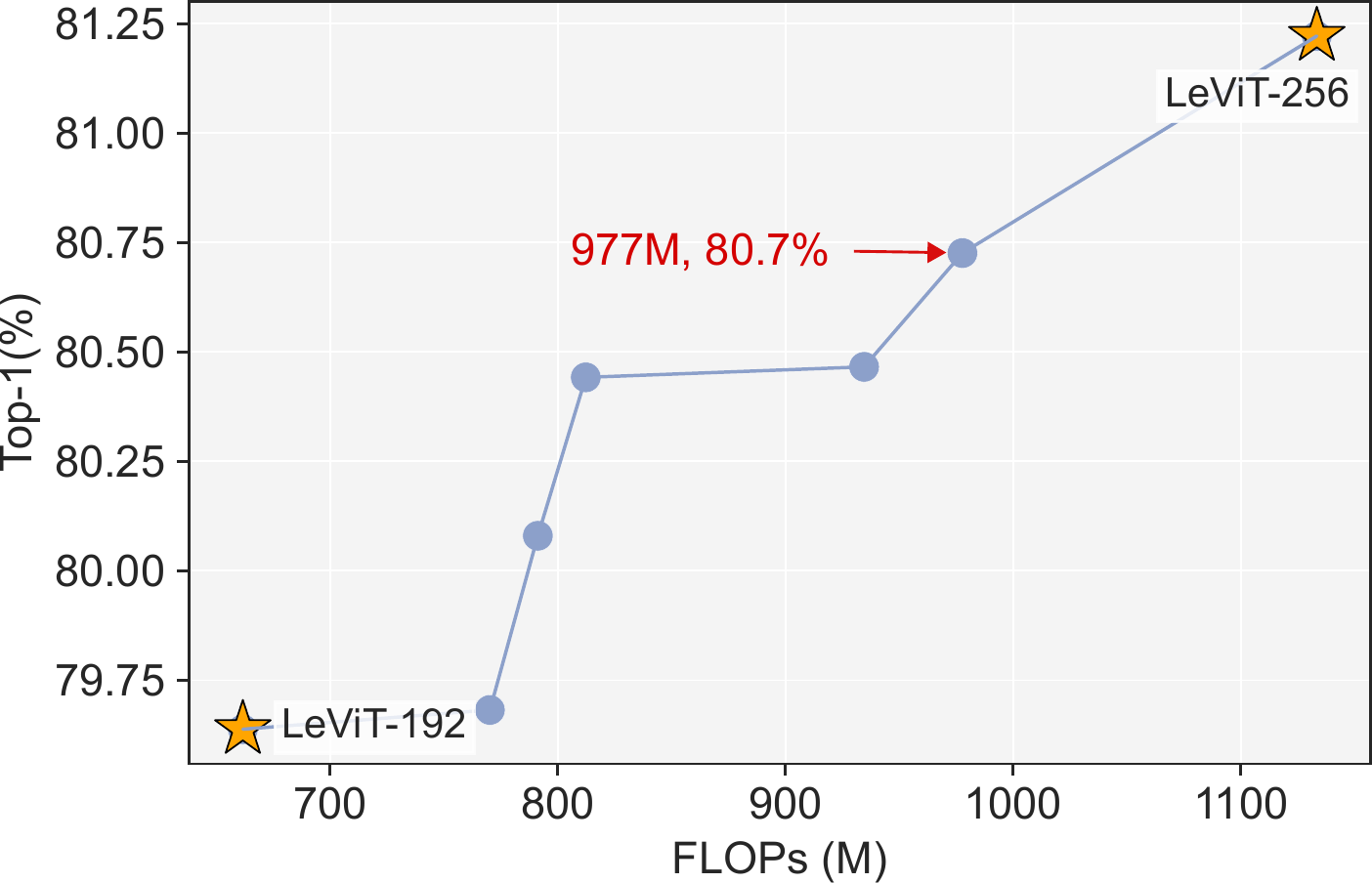}
	\caption{Stitching LeViT-192 and LeViT-256}
	\label{fig:stitch_levit}
\end{figure}

As discussed earlier, SN-Net is fundamentally different from one-shot NAS. Specifically, one-shot NAS trains a supernet from scratch and searches for an optimal sub-network during deployment to meet a specific resource constraint with complicated techniques (\eg, evolutionary search) and expensive cost (\eg, $>2K$ GPU hours in \cite{bignas}). In contrast, SN-Net aims to cheaply and fast assemble pretrained model families (\eg, $\sim$110 GPU hours) to get a scalable network, and instantly select optimal stitches due to the interpolation effect. In our experiments, we use DeiTs and Swins as two examples to show that SN-Net is a universal framework. Besides, we show in Figure~\ref{fig:stitch_levit} that we easily achieve comparable performance with BigNASModel-XL~\cite{bignas} (80.7\% \vs 80.9\%) with lower FLOPs (977M \vs 1040M) by stitching LeViTs~\cite{levit}.

\section{Compared with LayerDrop at Inference Time} \label{sec:compare_layerdrop}
LayerDrop~\cite{layerdrop} is a form of structured dropout which randomly drops Transformer layers during training for regularization. It also facilitates efficient pruning by dropping some layers at inference time. In DeiT-based SN-Net, the anchors are already pretrained with a drop rate of $0.1$. To show the advantage of our method, we train DeiT-B (\ie, the largest model in DeiT family) with a more aggressive path drop rate (0.5) and achieve 81.4\% Top-1 accuracy on ImageNet. However, cropping some layers of this trained network during testing performs badly, \eg, throwing the first 6 blocks (achieving 0.2\%), the last 6 blocks (52.7\%), and every other (72.7\% with 8.9G FLOPs), while our method achieves 72.6\% with 2.1G FLOPs.

%%%%%%%%% REFERENCES
{\small
\bibliographystyle{ieee_fullname}
\bibliography{egbib}
}

\end{document}